\relax
\documentclass[letterpaper]{article} % DO NOT CHANGE THIS
\usepackage{aaai21}  % DO NOT CHANGE THIS
\usepackage{times}  % DO NOT CHANGE THIS
\usepackage{helvet} % DO NOT CHANGE THIS
\usepackage{courier}  % DO NOT CHANGE THIS
\usepackage[hyphens]{url}  % DO NOT CHANGE THIS
\usepackage{graphicx} % DO NOT CHANGE THIS
\usepackage{natbib}  % DO NOT CHANGE THIS AND DO NOT ADD ANY OPTIONS TO IT
\usepackage{caption} % DO NOT CHANGE THIS AND DO NOT ADD ANY OPTIONS TO IT
\usepackage{multirow}
\usepackage{booktabs}
\usepackage{bm}
\usepackage{amsmath,amsthm,mathtools,amsfonts,amssymb}
\usepackage{graphicx,subcaption}
\usepackage{algorithm, algorithmic}

\title{OpEvo: An Evolutionary Method for Tensor Operator Optimization}
\author {
    Xiaotian Gao,
    Wei Cui,
    Lintao Zhang,
    Mao Yang \\
}
\affiliations {
    Microsoft Research \\
    {xiaotian.gao, weicu, lintaoz, maoyang}@microsoft.com
}

\begin{document}

\maketitle

\begin{abstract}
Training and inference efficiency of deep neural networks
highly rely on the performance of tensor operators on hardware platforms.
Manually optimizing tensor operators
has limitations in terms of supporting new operators or hardware platforms.
Therefore, automatically optimizing device code configurations of tensor operators is getting increasingly attractive.
However, current methods for tensor operator optimization usually suffer from poor sample-efficiency
due to the combinatorial search space.
In this work, we propose a novel evolutionary method, OpEvo,
which efficiently explores the search spaces of tensor operators
by introducing a topology-aware mutation operation based on q-random walk
to leverage the topological structures over the search spaces.
Our comprehensive experiment results show that
compared with state-of-the-art (SOTA) methods
OpEvo can find the best configuration with the lowest variance
and least efforts in the number of trials and wall-clock time.
All code of this work is available online.
\end{abstract}

\section{Introduction}\label{sec:introduction}

Abundant applications raise the demands of training and inference deep neural networks (DNNs) efficiently
on diverse hardware platforms ranging from cloud servers to embedded devices.
Moreover, computational graph-level optimization of deep neural network,
like tensor operator fusion~\cite{wang2010kernel}, may introduce new tensor operators.
Thus, manually optimized tensor operators provided by hardware-specific libraries have limitations in terms of
supporting new operators or hardware platforms,
so automatically optimizing tensor operators on diverse hardware platforms
is essential for large-scale deployment and application of deep learning technologies in the real-world problems.

Tensor operator optimization is essentially a combinatorial optimization problem.
The objective function is the performance of a tensor operator on specific hardware platform,
which should be maximized with respect to the hyper-parameters of corresponding device code,
such as how to tile a matrix or whether to unroll a loop.
Thereafter, we will refer to a tuple of hyper-parameters determining device code as a configuration,
and the set of all possible configurations as a configuration space or search space.
Unlike many typical problems of this type, such as travelling salesman problem,
the objective function of tensor operator optimization is a black box and expensive to sample.
One has to compile a device code with a specific configuration
and run it on real hardware to get the corresponding performance metric.
Therefore, a desired method for optimizing tensor operators should find the best configuration with as few samples as possible.

The expensive objective function makes solving tensor operator optimization problem with traditional combinatorial optimization methods,
for example, simulated annealing (SA)~\cite{kirkpatrick1983optimization} and evolutionary algorithms (EA)~\cite{back1993overview}, almost impossible.
Although these algorithms inherently support combinatorial search spaces~\cite{youssef2001evolutionary},
they do not take sample-efficiency into account,
thus thousands of or even more samples are usually needed,
which is unacceptable when tuning tensor operators in product environments.
On the other hand, sequential model based optimization (SMBO) methods
are proved sample-efficient for optimizing black-box functions with continuous search spaces~\cite{srinivas2009gaussian,hernandez2014predictive,wang2017max}.
However, when optimizing ones with combinatorial search spaces,
SMBO methods are not as sample-efficient as their continuous counterparts~\cite{hutter2011sequential},
because there is lack of prior assumptions about the objective functions,
such as continuity and differentiability in the case of continuous search spaces.
For example, if one could assume an objective function with a continuous search space is infinitely differentiable,
a Gaussian process with a radial basis function (RBF) kernel could be used to model the objective function.
In this way, a sample provides not only a single value at a point
but also the local properties of the objective function in its neighborhood or even global properties,
which results in a high sample-efficiency.
In contrast, SMBO methods for combinatorial optimization suffer from poor sample-efficiency
due to the lack of proper prior assumptions and corresponding surrogate models.

Besides sample-efficiency,
another weakness of SMBO methods is the extra burden introduced by training and optimizing surrogate models.
Although it can be safely ignored for many ultra-expensive objective functions,
such as hyperparameter tuning and architecture search for neural networks~\cite{elsken2018neural},
in which a trial usually needs several hours or more,
but it is not the case in the context of tensor operator optimization,
since compiling and executing a tensor operator usually need at most tens of seconds.

In this work, we propose a lightweight model-free method, OpEvo (\textbf{Op}erator \textbf{Evo}lution), which combines both advantages of EA and SMBO
by leveraging prior assumptions on combinatorial objective functions in an evolutionary framework.
Although there is no nice property like continuity or differentiability,
we construct topological structures over search spaces of tensor operators
by assuming similar configurations of a tensor operator will result in similar performance,
and then introduce a topology-aware mutation operation by proposing a $q$-random walk distribution
to leverage the constructed topological structures for better trade-off between exploration and exploitation.
In this way, OpEvo not only inherits the support of combinatorial search spaces and model-free nature of EA,
but also benefits from the prior assumptions about combinatorial objective functions,
so that OpEvo can efficiently optimize tensor operators. The contributions of the paper are four-fold:

\begin{itemize}
  \item We construct topological structures for search spaces of tensor operator optimization
  by assuming similar configurations of a tensor operator will result in similar performance;
  \item We define $q$-random walk distributions over combinatorial search spaces equipped with topological structures
  for better trade-off between exploitation and exploration;
  \item We propose OpEvo, which can leverage the topological structures over search spaces
  by introducing a novel topology-aware mutation operation based on $q$-random walk distributions;
  \item We evaluate the proposed algorithm with comprehensive experiments on both Nvidia and AMD platforms.
  Our experiments demonstrate that compared with state-of-the-art (SOTA) methods
  OpEvo can find the best configuration with the lowest variance
  and least efforts in the number of trials and wall-clock time.
\end{itemize}

The rest of this paper is organized as follows.
We summarize the related work in Section~\ref{sec:related work},
and then introduce a formal description of tensor optimization problem
and construct topological structures in Section~\ref{sec:problem formulation}.
In Section~\ref{sec:methodology},
we describe OpEvo method in detail and demonstrate its strength with experiments of optimizing typical tensor operators in Section~\ref{sec:experiments}.
Finally, we conclude in Section~\ref{sec:conclusion}.

\section{Related Work}\label{sec:related work}

As a class of popular methods for expensive black-box optimization,
SMBO methods are potential solutions for tensor operator optimization.
Although classic SMBO methods, such as Bayesian optimization (BO) with Gaussian process surrogate,
are usually used to optimize black-box functions with continuous search spaces,
many works have been done in using SMBO to optimize combinatorial black-box functions.
\citet{hutter2011sequential} proposed SMAC,
which uses random forest as a surrogate model to optimize algorithm configuration successfully.
\citet{bergstra2011algorithms} proposed TPE,
which uses tree-structured Parzen estimator as a surrogate model
to optimize hyperparameters of neural networks and deep belief networks.
As for tensor operator optimization,
TVM~\cite{chen2018tvm} framework implemented a SMBO method called AutoTVM~\cite{chen2018learning}
to optimize configurations of tensor operators.
Specifically, AutoTVM fits a surrogate model with either XGBoost~\cite{chen2016xgboost} or TreeGRU~\cite{tai2015improved},
and then uses SA to optimize the surrogate model for generating a batch of candidates in an $\epsilon$-greedy style.
\citet{ahn2020chameleon} proposed CHAMELEON to further improve AutoTVM with clustering based adaptive sampling and
reinforcement learning based adaptive exploration to reduce the number of costly hardware and surrogate model measurements, respectively.
Ansor~\cite{zheng2020ansor} is another work built upon TVM.
It used EA instead of SA to optimize surrogate models and devised an end-to-end framework to allocate computational resources among subgraphs and
hierarchically generate TVM templates for them.
Although these methods are successfully used in many combinatorial optimization problems,
they are not as sample-efficient as their continuous counterparts
due to the lack of proper prior assumptions and corresponding surrogate models.
OpEvo, on the other hand, introduces and leverages topological structures over combinatorial search spaces
thus obtains better sample and time-efficiency than previous arts.

AutoTVM and CHAMELEON also claimed that they are able to transfer knowledge among operators
through transfer learning and reinforcement learning, respectively.
However, they seem not so helpful in the context of tensor operator optimization.
For transfer learning, the pre-training dataset should be large and diverse enough to cover main information in fine-tuning datasets,
like ImageNet~\cite{deng2009imagenet} and GPT-3~\cite{brown2020language} did,
otherwise using a pre-trained model is more likely harmful than beneficial.
Many tensor operators needing optimizing are either new types of operators generated by tensor fusion or expected to run on new devices.
Neither is suitable for transfer learning.
Even if transfer learning works in some cases, pre-training a surrogate model with a large dataset before starting a new search and
executing and fine-tuning such model during searching are probably more time and money-consuming than just sampling more configurations on hardwares.
For reinforcement learning, its brittle convergence and poor generalization have been widely questioned for many years~\cite{haarnoja2018soft,cobbe2019quantifying,cobbe2019leveraging}.
There seems no guarantee that the policy learned by CHAMELEON can generalize to unseen operators so that improve sample-efficiency.
% Therefore, compared with current model-based methods, model-free methods cannot transfer knowledge among operators is not a weakness.

Two operator-specific methods,
Greedy Best First Search (G-BFS) and Neighborhood Actor Advantage Critic (N-A2C),
have been recently proposed to tune matrix tiling schemes of matrix multiplication (MatMul) operators
by taking the relation between different configurations into account~\cite{zhang2019compiler}.
They actually introduce a topology over the configuration space of MatMul operator by defining a neighborhood system on it,
and further employ a Markov Decision Process (MDP) for exploration over the configuration space.
By leveraging a domain-specific topological structure, G-BFS and N-A2C outperform AutoTVM in optimizing MatMul operators.
However, these two methods are only designed for tuning tiling schemes of multiplication of matrices with only power of 2 rows and columns,
so they are not compatible with other types of configuration spaces.
Further, they tend to encounter curse of dimensionality as the number of parameters needed tuning getting bigger,
because they only change one parameter at a time based on the MDP they defined.
Thus, generalizing them to more general tensor operators is not straightforward.
OpEvo, on the other hand, constructs topological structures in a general way and
uses evolutionary framework rather than MDP framework to explore search spaces,
so that the aforementioned problems encountered by G-BFS and N-A2C are overcame.

\section{Problem Formulation}\label{sec:problem formulation}

As earlier mentioned, tensor operator optimization is essentially a black-box optimization problem with a combinatorial search space.
It can be formally written as
\begin{equation}\label{eq:problem}
  \begin{gathered}
    x^\star=\underset{x\in\mathbb{X}}{\arg\max}\ f(x),\
    \mathbb{X}=\prod_{i=1}^\mu \mathcal{X}_i.
  \end{gathered}
\end{equation}
Here, $f(x)$ is a black-box function that measures the performance of a specific tensor operator with configuration $x$.
We use trillion floating-point operations per second (TFLOPS) as the measurement in this work.
Configuration $x$ is an ordered $\mu$-tuple $(x_1,...,x_\mu)$ and
each component $x_i\in\mathcal{X}_i$ corresponds to a hyperparameter of a device code,
so the entire search space $\mathbb{X}$ is the Cartesian product of all component feasible sets $\prod_{i=1}^\mu\mathcal{X}_i$.
Our aim is to find the optimal configuration $x^\star\in\mathbb{X}$ that corresponds to the maximum TFLOPS.

A topological structure over each $\mathcal{X}_i$ can be introduced by defining an undirected graph $G=(V,E)$,
% where the set of vertices $V\in\{\mathcal{X}_i|i=1,...,\mu\}$ is one of the component feasible sets,
where the set of vertices $V$ is $\mathcal{X}_i$,
and the set of edges $E=\{\{u,v\}|u,v\in V,\ u\neq v,\ g_V(u,v)=1\}$.
Here $g_V(u,v)$ is an adjacency function mapping from $V\times V$ to $\{0,1\}$.
$g_V(u,v)=1$ represents vertices $u$ and $v$ are adjacent, otherwise $u$ and $v$ are not adjacent.
In this way, one can introduce a topological structure over $V$ by defining an adjacency function $g_V(u,v)$
according to prior assumptions on $V$.
For example, it is intuitive to treat $u$ and $v$ as adjacent
if similar performance can be expected when changing from $u$ to $v$.
Search process can benefit from the topological structures introduced this way by obtaining information about neighborhood of samples,
like the performance of configurations in the neighborhood of a poor performance configuration are probably poor as well,
so that better sample-efficiency could be achieved.

Different tensor operators may have different types of hyperparameters and corresponding feasible sets.
In the rest part of this section,
we will discuss four kinds of hyperparameters used in this work,
and construct topological structures for them.
It should be noted that, besides them, one can easily introduce other types of hyperparameters
and construct corresponding topological structures based on concrete demands in a similar way.

\begin{figure}[ht]
  \centering
  \begin{subfigure}[h!]{0.23\textwidth}
    \centering
    \includegraphics[width=\textwidth]{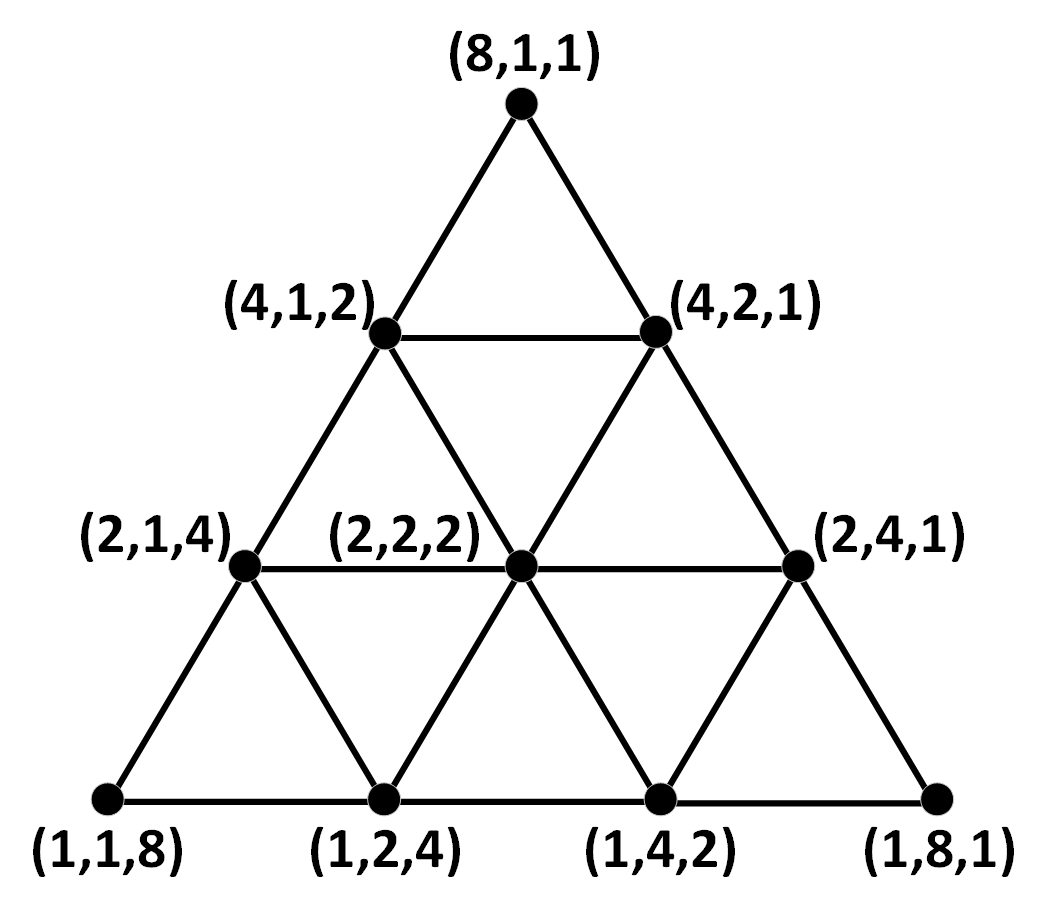}
    \caption{Factorization parameter with $C=8$ and $\nu=3$.}
    \label{fig:factor example}
  \end{subfigure}
  \begin{subfigure}[h!]{0.23\textwidth}
    \centering
    \includegraphics[width=\textwidth]{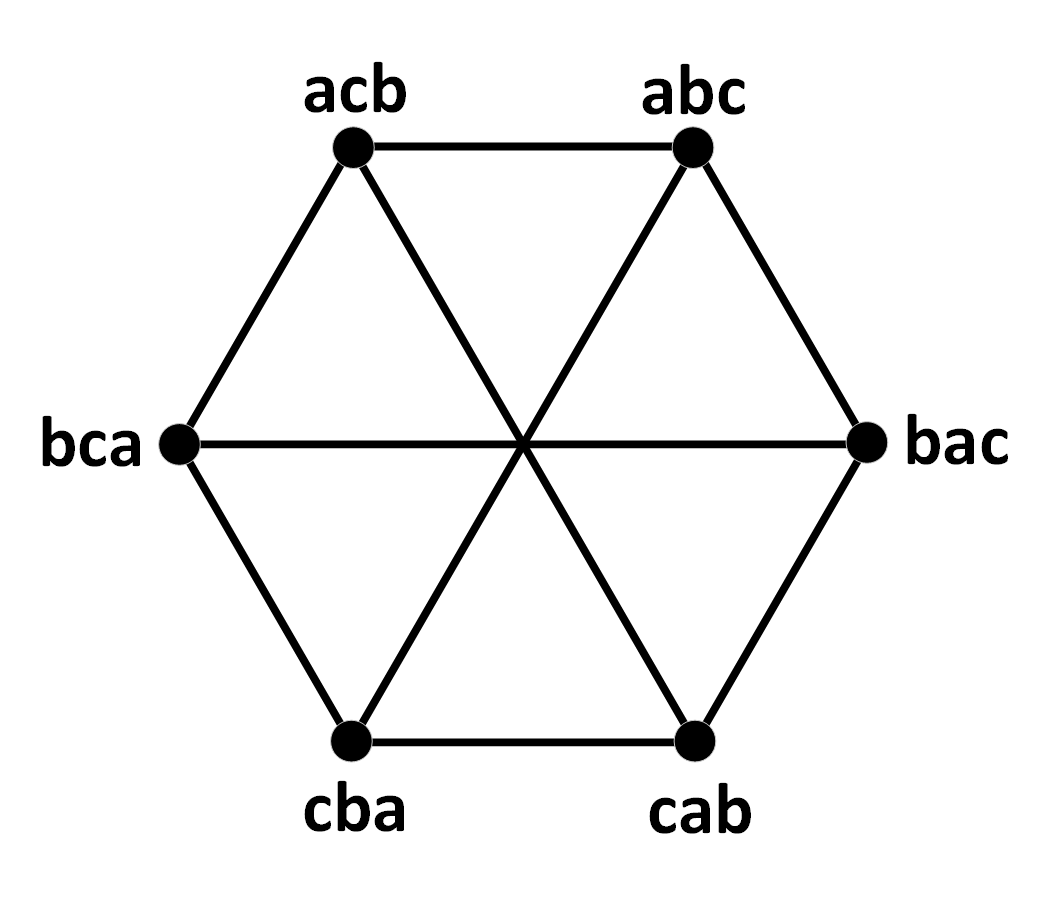}
    \caption{Permutation parameter with $n=3$.}
    \label{fig:perm example}
  \end{subfigure}
  \begin{subfigure}[h!]{0.23\textwidth}
    \centering
    \includegraphics[width=\textwidth]{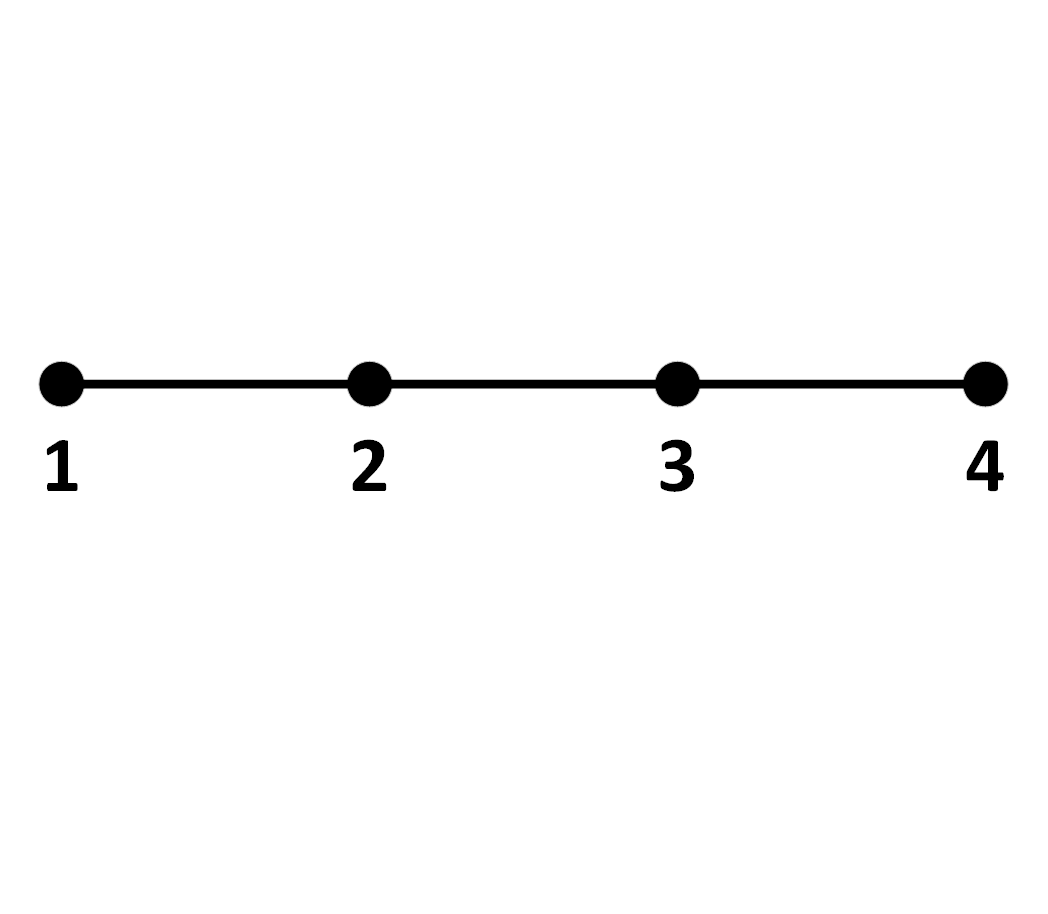}
    \caption{Discrete parameter with feasible set $\{1,2,3,4\}$.}
    \label{fig:discrete example}
  \end{subfigure}
  \begin{subfigure}[h!]{0.23\textwidth}
    \centering
    \includegraphics[width=\textwidth]{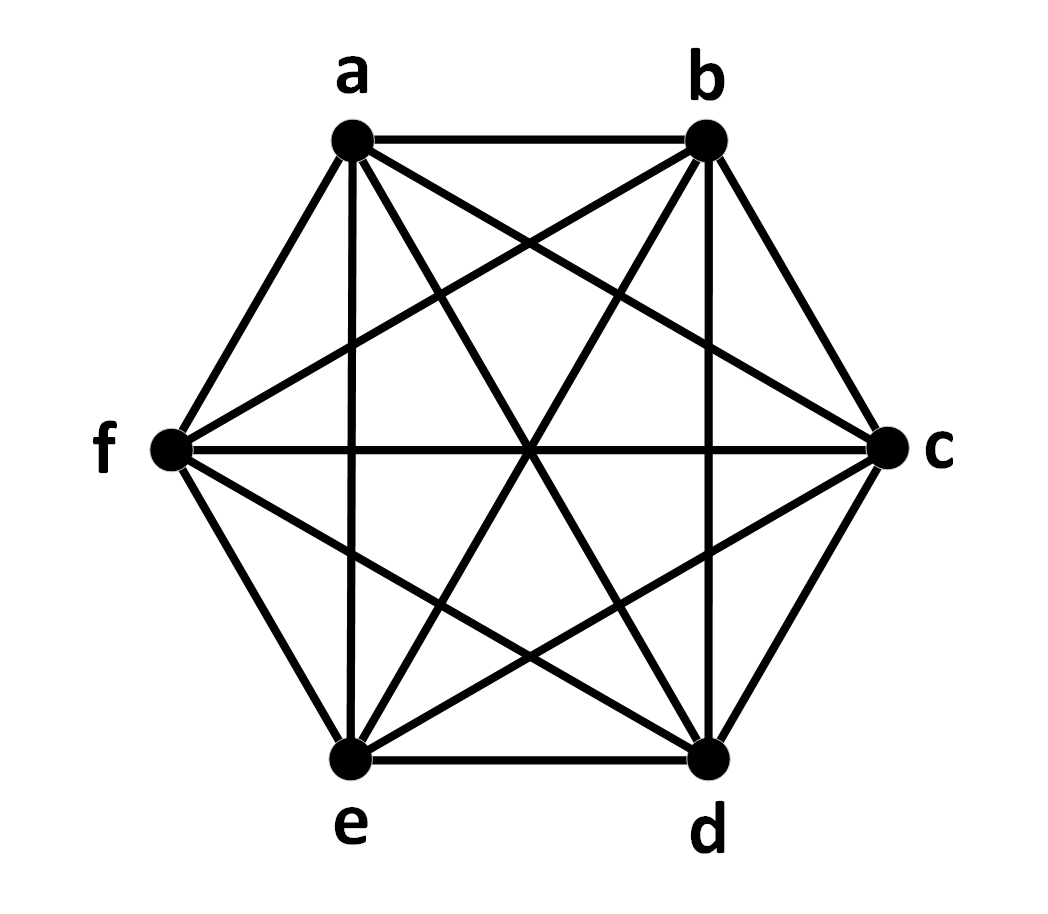}
    \caption{Categorical parameter with feasible set $\{a,b,c,d,e,f\}$.}
    \label{fig:catagorical example}
  \end{subfigure}
  \caption{A simple illustration of topological structures introduced over the search spaces of tensor operators.}
  \label{fig:topology example}
\end{figure}

First is the $\nu$-tuple with a factorization constraint,
$\mathcal{X}_i=\{(a_j)_{j=1}^\nu|\prod_{j=1}^\nu a_j=C,\ a_j\in\mathbb{N}_+\}$,
where $\nu,C\in\mathbb{N}_+$ are constants depending on specific tensor operators.
We will refer to this type of parameter as factorization parameter thereafter.
The factorization parameter is required by a popular technique called matrix tiling for improving the cache hit rate of memory access.
It iteratively splits computation into smaller tiles to adapt memory access patterns to a particular hardware.
From the implementation perspective, it transforms a single loop into nested loops,
where $\nu$ is the number of nested loops, $C$ is the total loop length and $a_j$ is the loop length of each nested loop.
We define two factorizations of $C$ are adjacent if one of them can be transformed to the other
by moving $z$, a prime factor of $C$, from the $n$-th factor to the $m$-th factor,
which is a basic transformation of the tiling scheme.
This adjacency function can be formally written as
$g(u,v)=1$ if $\exists n,m,z$ such that $u_m = v_m\cdot z$ and $u_n = v_n\cdot z^{-1}$, and $0$ otherwise,
where $n,m=1,...,\nu$ are distinct indices.
A simple example of the topology defined this way with $C=8$ and $\nu=3$
is illustrated in Figure~\ref{fig:factor example}.

The second type is the permutation parameter, $\mathcal{X}_i=\mathcal{M}!$,
where $\mathcal{M}$ is a set with $n$ distinct elements
and $\mathcal{M}!$ represents the symmetric group over $\mathcal{M}$.
The order of nested loops in device code can be modeled by this type of parameter,
where $n$ is the number of nested loops and each element in the feasible set corresponds to a particular order of nested loops.
We define two permutations of $\mathcal{M}$ are adjacent if one of them can be transformed to the other by a two-cycle permutation,
which is a basic transformation of the order.
This adjacency function can be formally written as
$g(u,v)=1$ if there exists a two-cycle permutation $\sigma$ of $\mathcal{M}$ such that $u=\sigma v$, and $0$ otherwise.
Figure~\ref{fig:perm example} shows the topology defined this way when $n=3$.

The third type is the discrete parameter,
$\mathcal{X}_i=\{a_j|j=1,...,J\ \mathrm{and}\ J\in\mathbb{N}_+,\ a_j\in\mathbb{R}\}$,
in which there are finite numbers.
The maximum step of loop unrolling is an example of discrete type parameter.
There is a natural adjacency among discrete parameters since they have well-defined comparability.
This natural adjacency function can be formally written as $g(u,v)=1$ if $\nexists w\in V$ such that $(w-u)\cdot(w-v)<0$, and $0$ otherwise.
A simple example of the topology defined this way with $\mathcal{X}_i=\{1, 2, 3, 4\}$
is illustrated in Figure~\ref{fig:discrete example}.

The last type is the categorical parameter,
$\mathcal{X}_i=\{a_j|j=1,...,J\ \mathrm{and}\ J\in\mathbb{N}_+\}$,
in which there are finite elements that can be any entity.
The choices like whether to unroll a loop and which thread axis to dispatch computation are examples of categorical type parameter.
Unlike discrete parameters, there is no natural adjacency among categorical parameters,
so all elements in the feasible set of categorical parameter are treated as adjacent,
which can be formally written as $g(u,v)=1$ for all $u,v\in V$ and $u\neq v$, and $0$ otherwise.
A simple example of the topology defined this way with $\mathcal{X}_i=\{a, b, c, d, e, f\}$
is illustrated in Figure~\ref{fig:catagorical example}.

\section{Methodology}\label{sec:methodology}

\subsection{Evolutionary Algorithm}

EA is a kind of stochastic derivative-free optimization methods,
which can be used to solve problems defined by Equation~\ref{eq:problem}.
EA imitates the natural selection in the evolution process of biological species
to find the optimal configuration of an objective function.
Evolutionary concepts are translated into algorithmic operations,
i.e., selection, recombination, and mutation~\cite{kramer2016machine},
which significantly influence the effectiveness and efficiency of EA.

To efficiently search the best configuration of a tensor operator,
OpEvo leverages topological structures defined in Section~\ref{sec:problem formulation} with an evolutionary framework.
In specific, OpEvo evolves a population of configurations, which are also called individuals in EA terminology.
The TFLOPS of executing tensor operators on a target hardware is a measure of the individuals' quality or fitness.
At each evolutionary step, we select top ranked individuals to be parents based on their fitnesses,
and then recombine and mutate them to generate new individuals or children.
After evaluation, children are added to the population to be candidates of new parents at the next evolutionary step.
This iteration will repeat until some termination criteria are met.

In the rest of this section, we will describe the selection, recombination and mutation operations of OpEvo in detail
and illustrate how OpEvo leverages the topological structures and why OpEvo can outperform previous arts in this way.

\subsection{Selection and Recombination}\label{sec:select}

Suppose we already have a list of individuals which are ranked by their fitnesses.
Top-$\lambda$ ranked individuals are chosen to be parents,
where $\lambda\in\mathbb{N}_+$ governs the diversity in evolutionary process.
Evolution with large $\lambda$ tends to get rid of suboptimum but sacrifices data efficiency,
while one with small $\lambda$ is easier to converge but suffers from suboptimum.

A child will be generated by recombining these selected parents in a stochastic way.
Specifically, we sample below categorical distribution with $\lambda$ categories $\mu$ times
to decide which parents each parameter of a child should inherit from.
\begin{equation}\label{eq:recombine}
  \begin{gathered}
    P(x_i = x_i^j) = \frac{f(x^j)}{\sum_{k=1}^\lambda f(x^k)}, \\
    \mathrm{for}\quad i=1,...,\mu,\quad j=1,...,\lambda,
  \end{gathered}
\end{equation}
where $\mu$ is the number of parameters in a configuration,
superscripts represent different parents, and subscripts represent different parameters in a configuration.
$x_i$ is the $i$-th parameter of generated child $x$.

% Equation~\ref{eq:recombine} is a fitness-related distribution
% in which a parent with larger fitness has bigger probability to transfer its characters to offspring.
% Rank-based fitness shaping is a popular technique
% often used to avoid suboptimum and accelerate convergence~\cite{hansen1996adapting,wierstra2008natural}.
% However, we don't use it because meaningful fitnesses are quite sparse when optimizing some tensor operators,
% rank-based fitness shaping is harmful rather than helpful in these cases.

It is worthwhile to mention that
many SOTA methods suffer invalid configurations in the search spaces,
which is inevitable since the constraints in search spaces are usually black-box as well.
OpEvo can mitigate this problem by assigning zero fitnesses to the invalid configurations
so that their characters have no chance to be inherited.
In this way, invalid configurations will have less and less probability to be sampled during evolution.

\subsection{Mutation}\label{sec:mutate}
% After selection and recombination, there is $\nu$
OpEvo mutates each parameter $x_i$ of each child
by sampling a topology-aware probability distribution over corresponding feasible set $\mathcal{X}_i$.
Given a topology over $\mathcal{X}_i$ and current vertex,
such topology-ware probability distribution can be constructed by a random walk-like process.
The transition probability from vertex $u$ to an adjacent vertex $v$ is
\begin{equation}\label{eq:dist}
  P_{uv}=\frac{q}{|S(u)|}, v\in S(u),
\end{equation}
where $q\in(0,1)$ is the mutation rate which trade-offs the exploration and exploitation.
OpEvo tends to exploration as $q$ approaches $1$, while tends to exploitation as $q$ approaches $0$.
$S(u)=\{v|g(u,v)=1\}$ is the set of all vertices adjacent to $u$,
and $|S(u)|$ denotes the cardinality of set $S(u)$.
The major difference between the "random walk" defined by Equation~\ref{eq:dist} and the regular random walk is that
the summation of transition probability over all adjacent vertices is $q$ rather than $1$,
so the "random walk" we introduced is not a Markov chain since there is a probability of $1-q$ to stop walking.
In this way, given a current vertex $u\in\mathcal{X}_i$,
the topology-aware probability distribution $P_u(v)$ for all $v\in\mathcal{X}_i$ could be defined as
the probability of walking started from $u$ and stopped at $v$.
We will refer to the distribution defined this way as $q$-random walk distribution thereafter.
Appendix~\ref{sec:proof} formally proved that the $q$-random walk distribution is a valid probability distribution over $\mathcal{X}_i$.

\begin{figure}[ht]
  \centering
  \begin{subfigure}[h!]{0.23\textwidth}
    \centering
    \includegraphics[width=\textwidth]{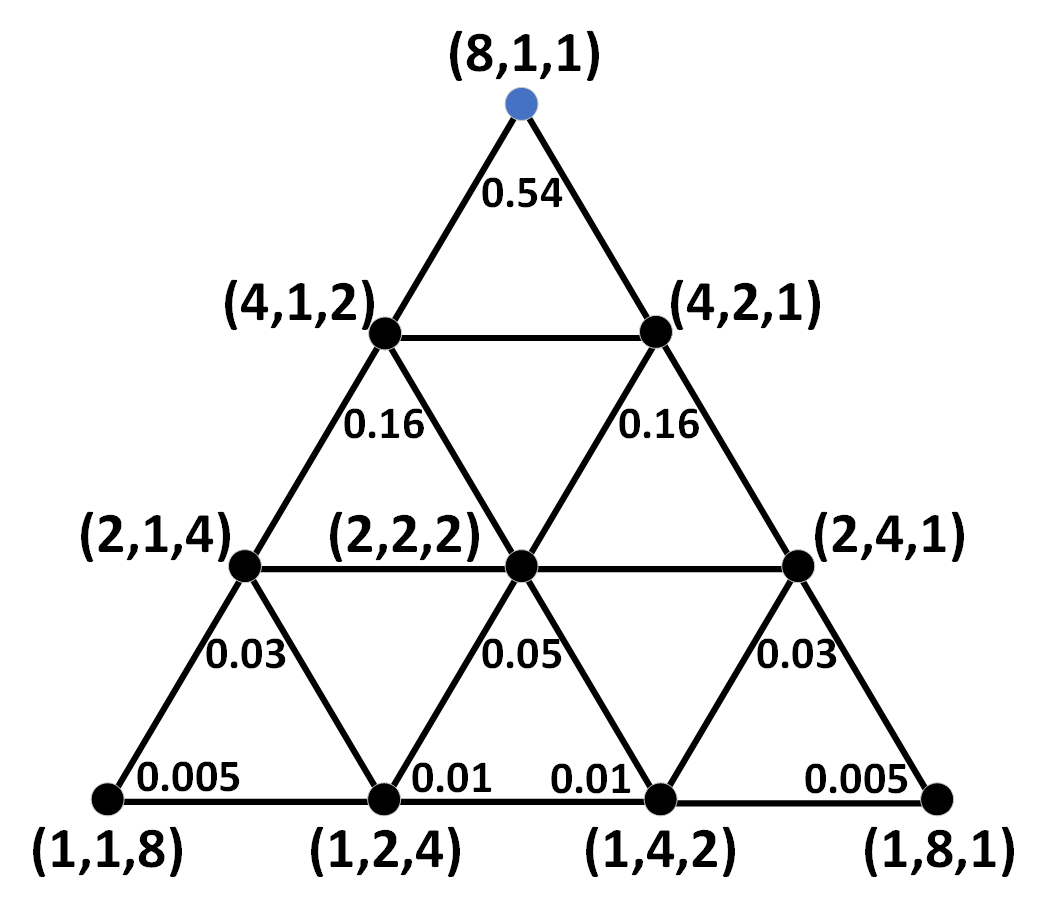}
    \caption{$q=0.5$}
    \label{fig:q example a}
  \end{subfigure}
  \begin{subfigure}[h!]{0.23\textwidth}
    \centering
    \includegraphics[width=\textwidth]{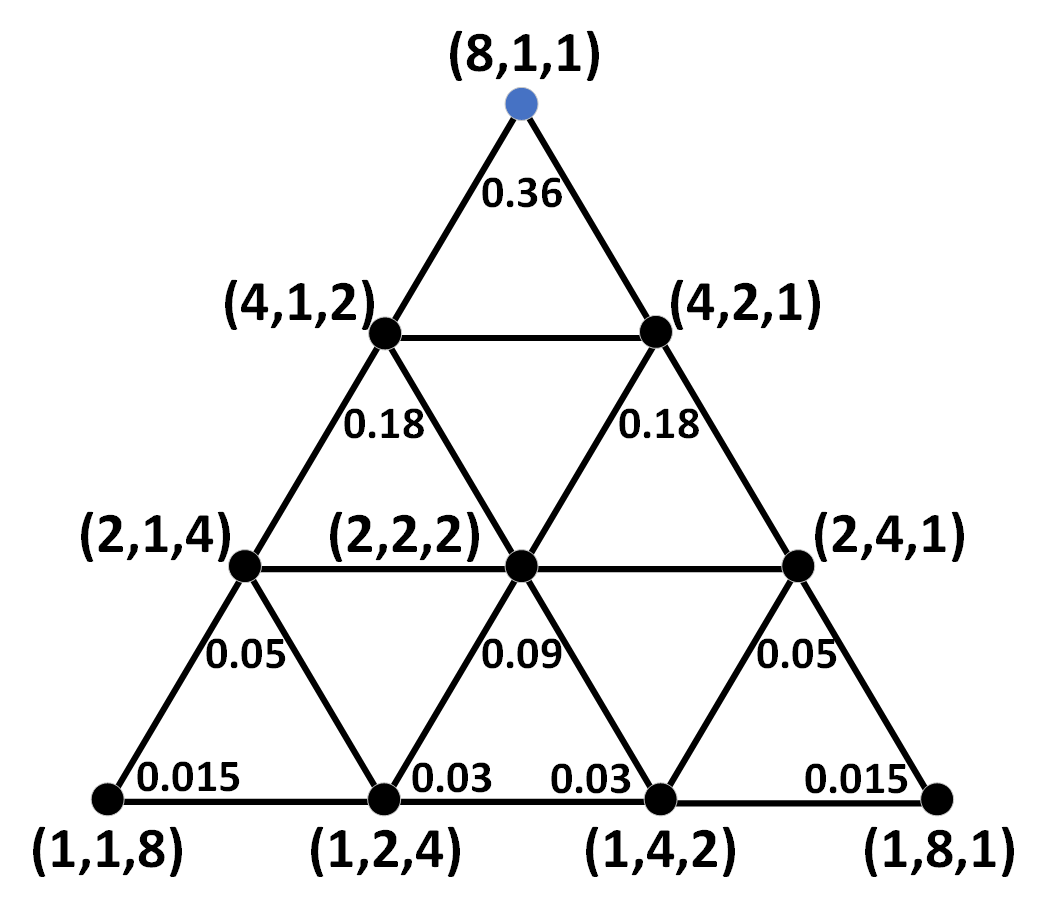}
    \caption{$q=0.7$}
    \label{fig:q example b}
  \end{subfigure}
  \caption{Two $q$-random walk distributions with different $q$.}
  \label{fig:q example}
\end{figure}

For revealing the intuition behind $q$-random walk distribution,
two q-random walk distributions over the feasible set of factorization parameter with $C=8$ and $\nu=3$
are illustrated in Figure~\ref{fig:q example}.
They start from the same vertex (the blue vertex) but mutate with different $q$.
It could be easily observed that the vertices with smaller distance to the start vertex have higher probability to be sampled,
which ensures a good trade-off between exploration and exploitation.
Further, the distribution with a larger $q$ has a wider spread than one with a smaller $q$,
because larger $q$ encourages more jumps in the $q$-random walk process.
Considering the asymptotic case of $q=1$, the $q$-random walk degenerates into a regular random walk on an undirected graph,
which keeps jumping forever and eventually traverses all vertices on the graph,
while in the case of $q=0$, the $q$-random walk vanishes and no mutation acts on parameter $x_i$.
Thus, $q$ is a hyperparameter for OpEvo to trade off exploitation and exploration.

Considering a regular random walk on an undirected graph, i.e. $q=1$,
the probability of visiting a vertex in the graph is determined by the graph topology
when the Markov chain induced by the random walk is converged.
That's why random walk can be used for embedding graphs in many works~\cite{perozzi2014deepwalk}.
$q$-random walk distribution also inherits this topology-aware nature.
Observing vertices with the same distance to the start vertex in Figure~\ref{fig:q example},
the vertices with more complex neighborhood have larger probability.
For example, vertices $(2,1,4)$, $(2,2,2)$ and $(2,4,1)$ have the same distance to start vertex $(8,1,1)$,
but vertex $(2,2,2)$ has larger probability since it has larger degree.
This property of $q$-random walk distribution helps explore search spaces efficiently,
because sampling vertices with more complex neighborhood will provide us more knowledge about objective functions.

\subsection{Summary}

The OpEvo algorithm is summarized in Algorithm~\ref{alg:opevo}.
At first, $\lambda$ configurations are randomly generated and evaluated
to initialize a priority queue $\mathcal{Q}$ ordered by decreasing fitness.
Next, taking top $\lambda$ configurations from $\mathcal{Q}$ as parents and
recombining them to generate $\rho$ children according to Section~\ref{sec:select}.
Then, each child is mutated based on Section~\ref{sec:mutate}.
Note that the same configuration will not be sampled twice in the whole process of OpEvo,
since the noise of TFLOPS of executing a tensor operator on a hardware is relatively small
and data efficiency can benefit from non-duplicated samples.
As a result, when a mutated child has already in $\mathcal{Q}$,
we will mutate the child again until it is not already sampled.
Finally, the fitnesses of $\rho$ children are evaluated on target hardware, and enqueued into $\mathcal{Q}$.
This iteration will repeat until some termination criteria are met.

\begin{algorithm}[htb]
  % \small
  \caption{OpEvo}
  \label{alg:opevo}
  \textbf{Input}: all component feasible sets $\mathcal{X}_i, i=1,..,\mu$,
  parents size $\lambda$, offspring size $\rho$, mutation rate $q$\\
  \textbf{Output}: optimal configuration $x^\star$
  \begin{algorithmic}[1] %[1] enables line numbers
    \STATE randomly generate $\lambda$ configurations $\{x^j\}_{j=1}^\lambda$
    \STATE evaluate $\{x^j\}_{j=1}^\lambda$ to get associated fitness,
           and enqueue $\{x^j,f(x^j)\}_{j=1}^\lambda$ into a priority queue $\mathcal{Q}$
    \REPEAT
    \STATE select $\lambda$ parents from $\mathcal{Q}$ and
    recombine them to generate $\rho$ children according to Section~\ref{sec:select}
    \STATE mutate $\rho$ children according to Section~\ref{sec:mutate}
    \STATE evaluate $\rho$ children on hardware, and enqueue $\{x^j,f(x^j)\}_{j=1}^\rho$ into $\mathcal{Q}$
    \UNTIL{termination criterion is met}
    \STATE \textbf{return} the best configuration so far
  \end{algorithmic}
\end{algorithm}

\section{Experiments}\label{sec:experiments}

We now evaluate the empirical performance of the proposed method with three typical kinds of tensor operators,
MatMul, BatchMatMul, 2D Convolution, and a classic CNN architecture AlexNet~\cite{krizhevsky2012imagenet} on both Nvidia (GTX 1080Ti) and AMD (MI50 GPU) platforms.
All tensor operators in our experiments are described and generated with TVM framework,
and then compiled and run with CUDA 10.0 or RCOM 2.9.
Additionally, we compare OpEvo with three aforementioned SOTA methods, G-BFS, N-A2C and AutoTVM.
In our experiments, OpEvo, G-BFS and N-A2C are implemented by ourselves
with the framework of Neural Network Intelligence (NNI, \textit{https://github.com/microsoft/nni/}),
and AutoTVM is implemented by its authors in the TVM project (\textit{https://github.com/apache/incubator-tvm}).
All codes for OpEvo, G-BFS, N-A2C and our benchmarks are publicly available with the NNI project.
Please refer to Appendix~\ref{sec:exp_details} for more details about the experiments
and Appendix~\ref{sec:omitted} for specific numbers about figures presented in this section.

\begin{figure}[ht]
  \centering
  \includegraphics[width=0.47\textwidth]{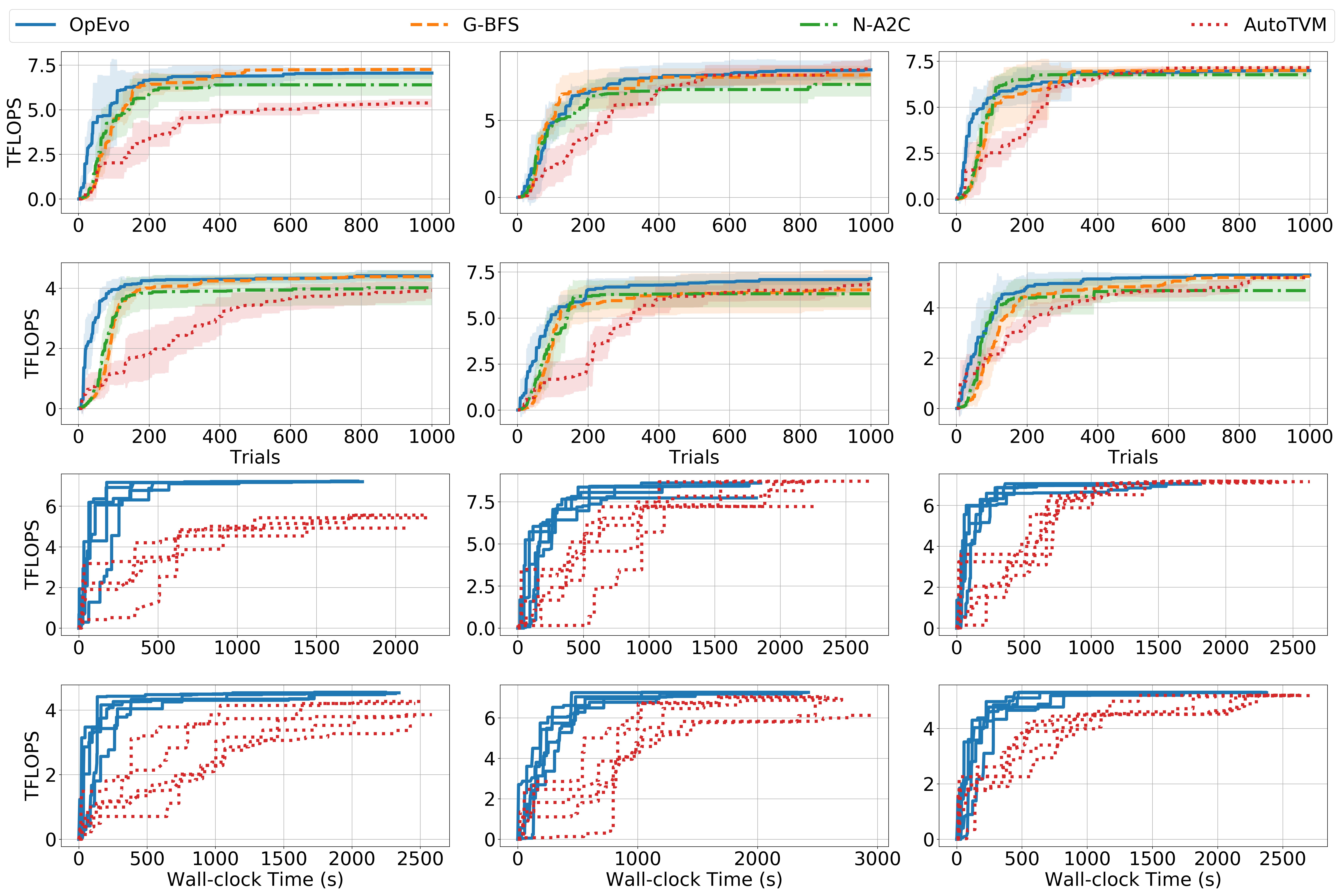}
  \caption{Algorithms comparison for three MatMul operators.
  The first and third rows are results on Nvidia platform, while the second and fourth rows are results on AMD platform.
  Three columns correspond to three operators MM1, MM2 and MM3 described in Appendix~\ref{sec:exp_mm} from left to right, respectively.}
  \label{fig:matmul}
\end{figure}

\subsection{MatMul}
Three different MatMul operators are chosen from BERT~\cite{devlin2018bert} to evaluate proposed method.
The maximum performance obtained so far versus number of trials and wall-clock time which have been used is illustrated in Figure~\ref{fig:matmul}.
For the upper two rows, the lines denote the averages of 5 runs, while the shaded areas indicate standard deviations.
For the lower two rows, each line denotes a specific run.
Different colors and line styles represent different algorithms.
From the results, it can be easily concluded that the methods leveraging predefined topology, OpEvo, G-BFS and N-A2C,
much outperform the general SMBO method, AutoTVM.
G-BFS and N-A2C leverage the underlying topology by introducing a MDP,
so just local topology is considered and leveraged to explore the configuration space,
while OpEvo can consider the global topology thanks to the mutation operation based on the $q$-random walk distribution.
Therefore, OpEvo performs better than G-BFS and N-A2C in most cases in terms of mean and variance of best TFLOPS and data-efficiency.
Further, as earlier mentioned, OpEvo is a lightweight model-free method,
so the extra burden for training and optimizing surrogate models is avoided.
It can be seen from Figure~\ref{fig:matmul} that OpEvo can save around $30\%$ and $10\%$ wall-clock time
when optimizing CUDA and ROCM operators, respectively.
This is because the CUDA compilation speed is usually faster than ROCM,
so the extra burden of tuning CUDA operators takes a larger share of the total wall-clock time.

\begin{figure}[ht]
  \centering
  \includegraphics[width=0.47\textwidth]{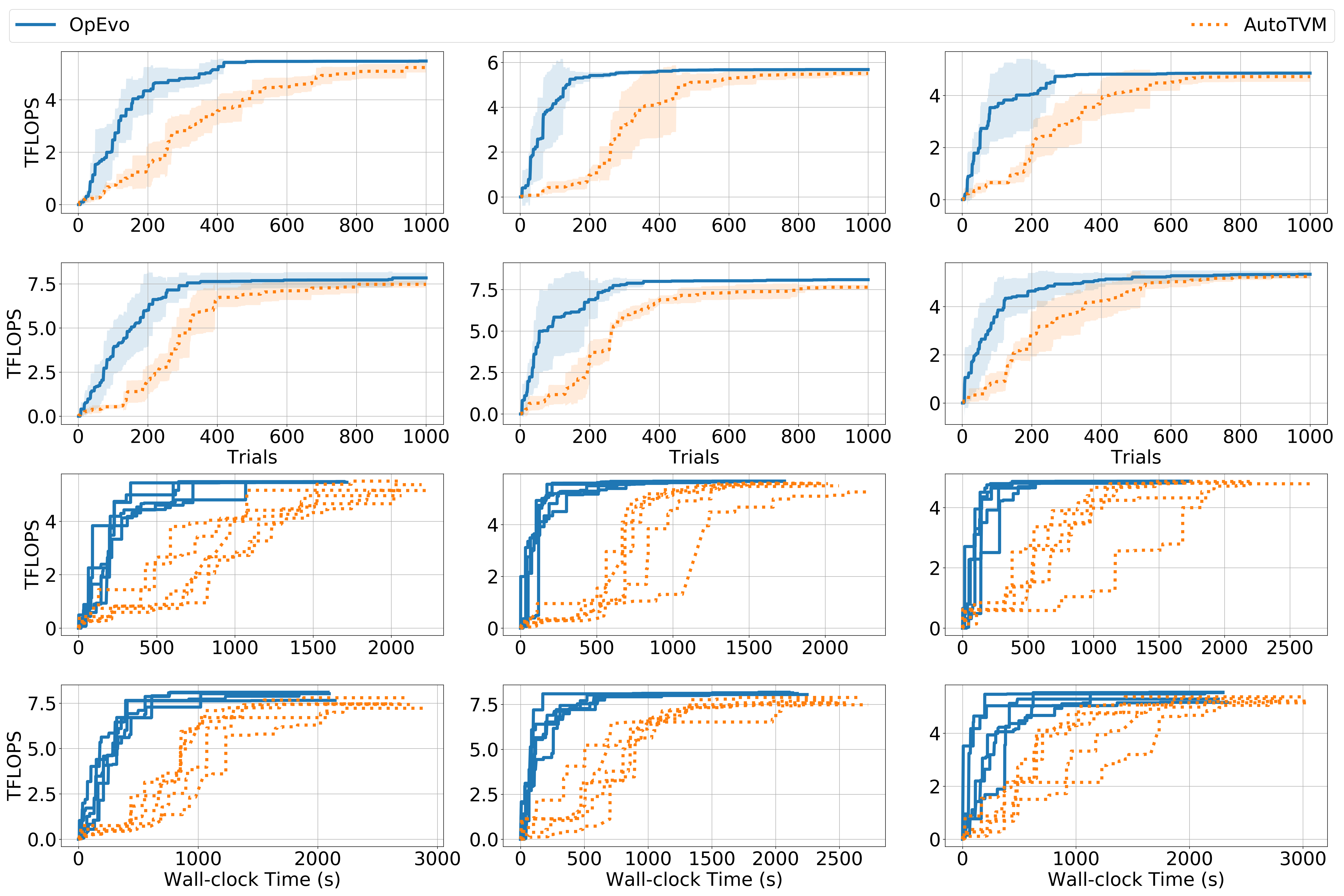}
  \caption{Algorithms comparison for three BatchMatMul operators.
  The first and third rows are results on Nvidia platform, while the second and forth rows are results on AMD platform.
  Three columns correspond to three operators BMM1, BMM2 and BMM3 described in Appendix~\ref{sec:exp_bmm} from left to right, respectively.}
  \label{fig:bmm}
\end{figure}

\begin{figure}[ht]
  \centering
  \includegraphics[width=0.314\textwidth]{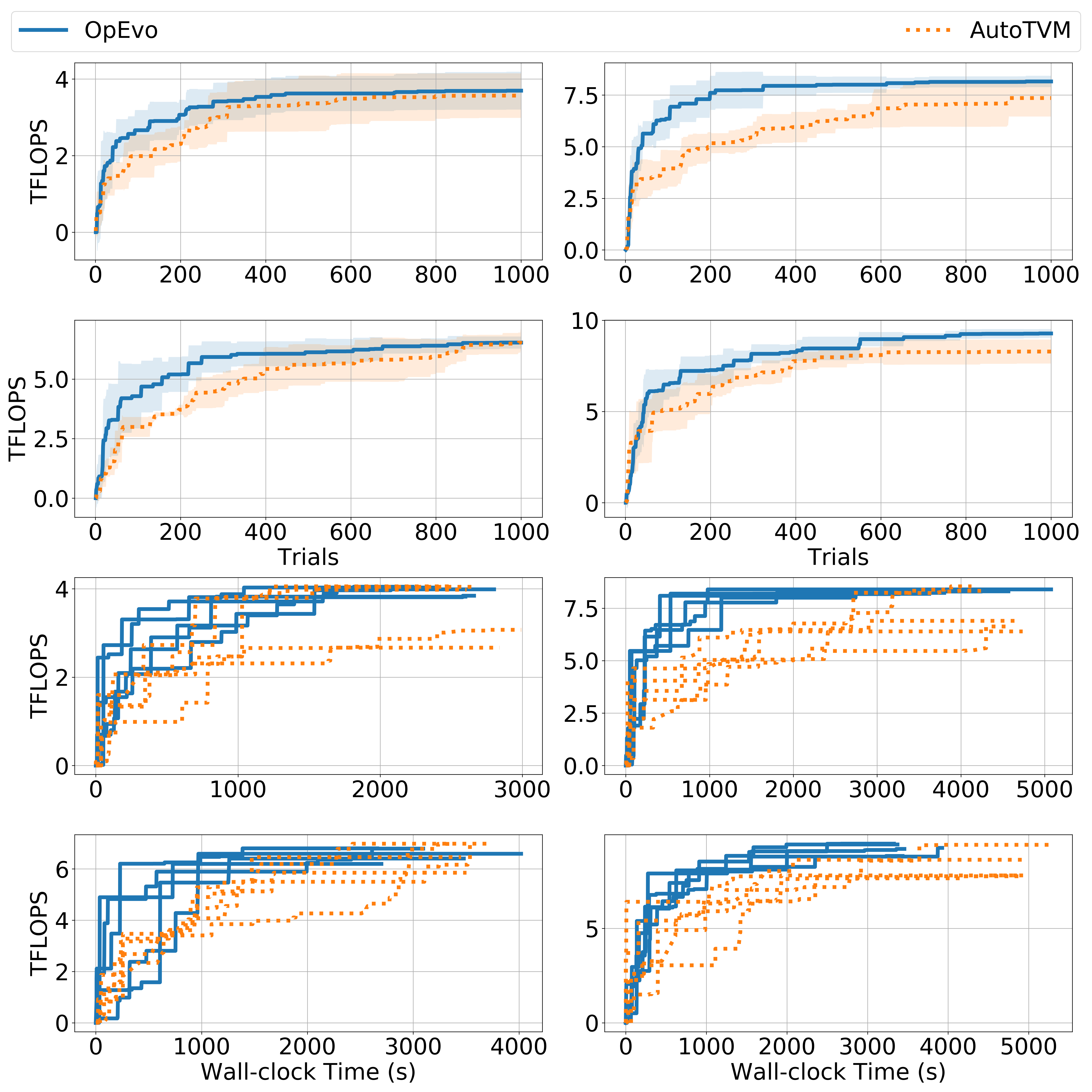}
  \caption{Algorithms comparison for two 2D Convolution operators.
  The first and third rows are results on Nvidia platform, while the second and forth rows are results on AMD platform.
  Two columns correspond to two operators C1 and C2 described in Appendix~\ref{sec:exp_conv} from left to right, respectively.}
  \label{fig:conv}
\end{figure}

\subsection{BatchMatMul}
There are also three BatchMatMul operators selected from BERT for evaluation.
All these operators have batch size $960$, so G-BFS and N-A2C are not capable to optimize them
because they can only deal with matrices with power of 2 rows and columns.
The comparison between OpEvo and AutoTVM is shown in Figure~\ref{fig:bmm}.
Compared with MatMul operators, BatchMatMul has an order of magnitude bigger search space
since one more parameter needed to be optimized.
Also, the generated BatchMatMul device code is more likely to overflow the device memory
as tile size of BatchMatMul is bigger than that of MatMul,
which leads to sparser performance measurement.
Although these challenges exist, OpEvo performs still well thanks to the globally exploration mechanism.
The variance of best performance even better than that of MatMul because of the sparsity.

\subsection{2D Convolution}
Two 2D convolution operators are chosen from AlexNet for evaluation.
They have more complex search spaces and thus harder to model compared with tensor operators discussed before,
since, besides factorization parameter, discrete and categorical parameters are also involved.
As a result, G-BFS and N-A2C are not capable to tune them.
Figure~\ref{fig:conv} shows the comparison between OpEvo and AutoTVM.
Although XGBoost is a tree boosting model
which is relatively friendly to discrete and categorical parameters,
AutoTVM still performs worse than OpEvo, because EA inherently supports complex search space
and OpEvo further improves sample-efficiency by leveraging predefined topology.
We note that the time-saving effect of OpEvo is not significant in the 2D convolution cases,
because compiling and executing convolution operators are much more time-consuming than MatMul and BatchMatMul Operators
and thus dominate the total tuning time.

\subsection{End-to-end Evaluation}

\begin{figure}[ht]
  \centering
  \includegraphics[width=0.47\textwidth]{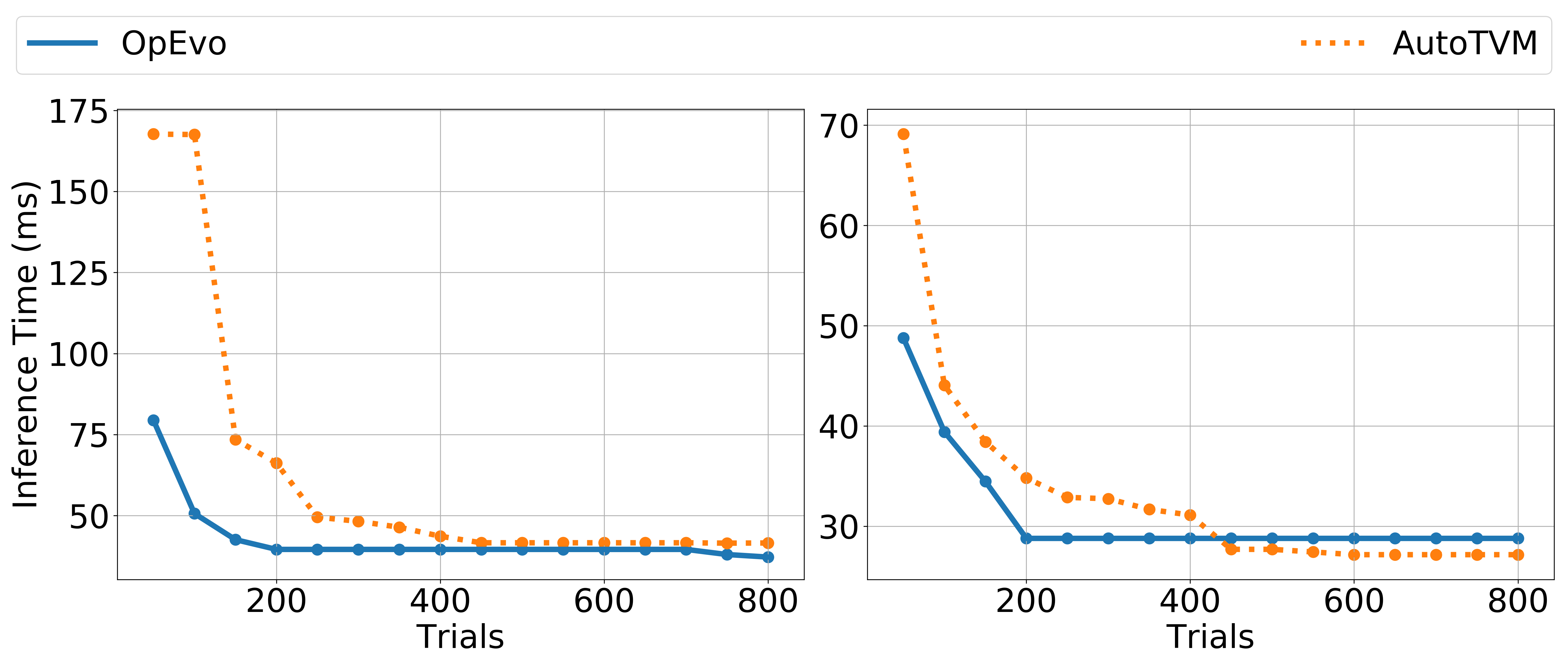}
  \caption{Algorithms comparison in terms of end-to-end inference time.
  The left figure is the result on Nvidia platform, while the right one is the result on AMD platform.}
  \label{fig:e2e}
\end{figure}

A classic CNN architecture, AlexNet, is used to evaluate the end-to-end performance of the proposed method,
where there are 26 different kinds of tensor operators covering the most commonly used types.
Figure~\ref{fig:e2e} shows the comparison between OpEvo and AutoTVM in terms of inference time,
form which it can be easily concluded that OpEvo is more data-efficient than AutoTVM on both NVIDIA and AMD platforms.
For OpEvo, the end-to-end inference time rapidly deceases and reaches the minimum at around 200 trials,
while AutoTVM needs at least 400 trails to reach the same performance.

\begin{figure}[ht]
  \centering
  \begin{subfigure}[h!]{0.47\textwidth}
    \centering
    \includegraphics[width=\textwidth]{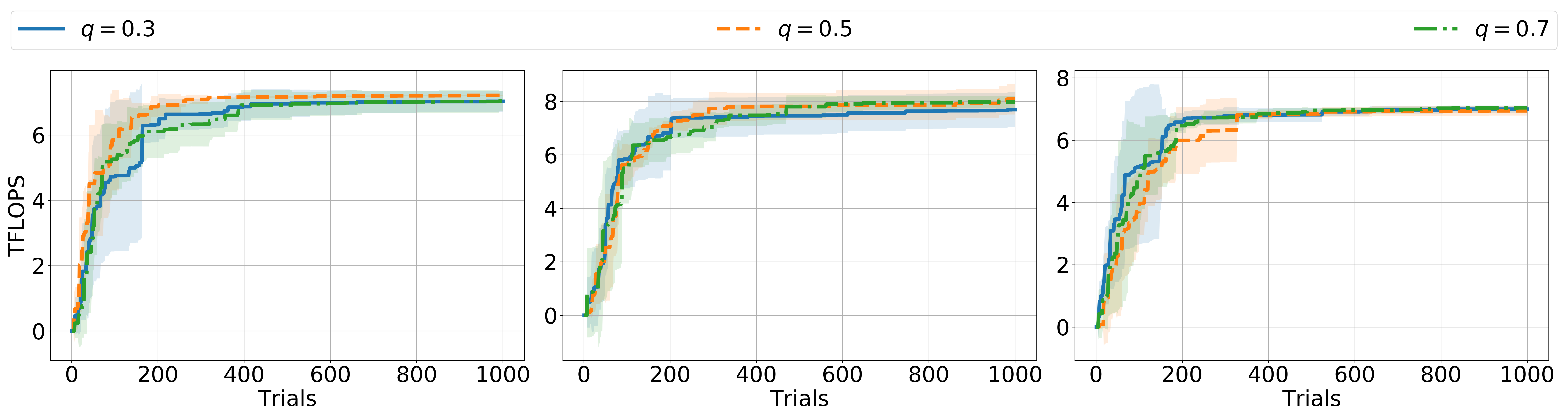}
  \end{subfigure}
  \begin{subfigure}[h!]{0.47\textwidth}
    \centering
    \includegraphics[width=\textwidth]{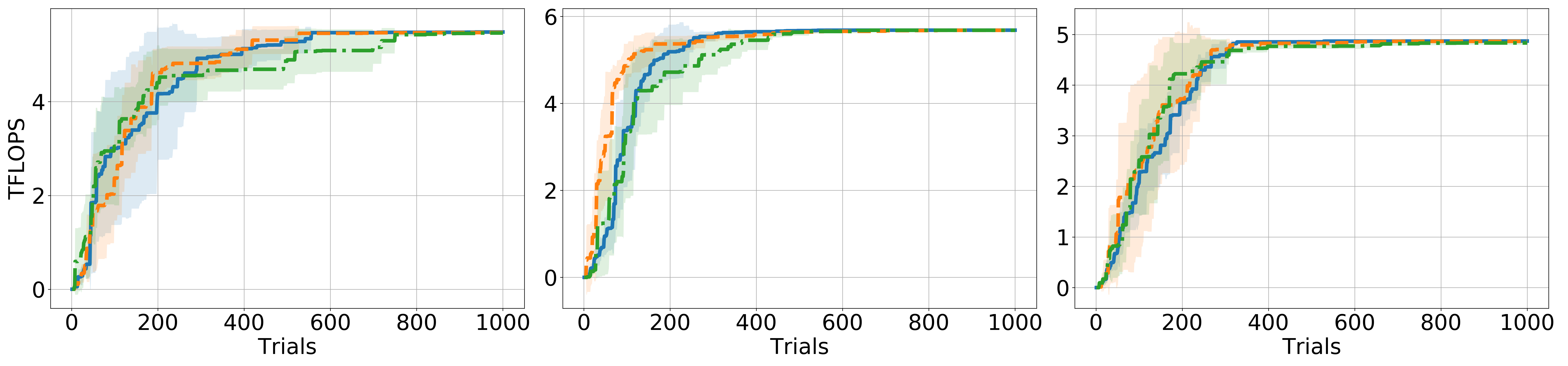}
  \end{subfigure}
  \begin{subfigure}[h!]{0.47\textwidth}
    \centering
    \includegraphics[width=\textwidth]{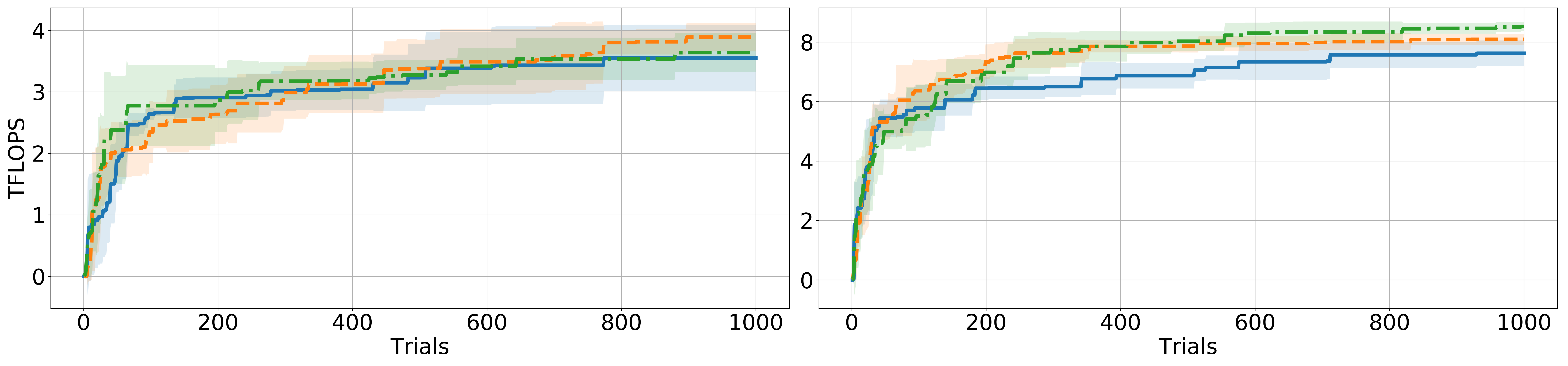}
  \end{subfigure}
  \caption{The effect of mutation rate $q$ on OpEvo.}
  \label{fig:q}
\end{figure}

\subsection{Hyperparameter Sensitivity}\label{sec:sensitivity}

\begin{figure}[ht]
  \centering
  \begin{subfigure}[h!]{0.47\textwidth}
    \centering
    \includegraphics[width=\textwidth]{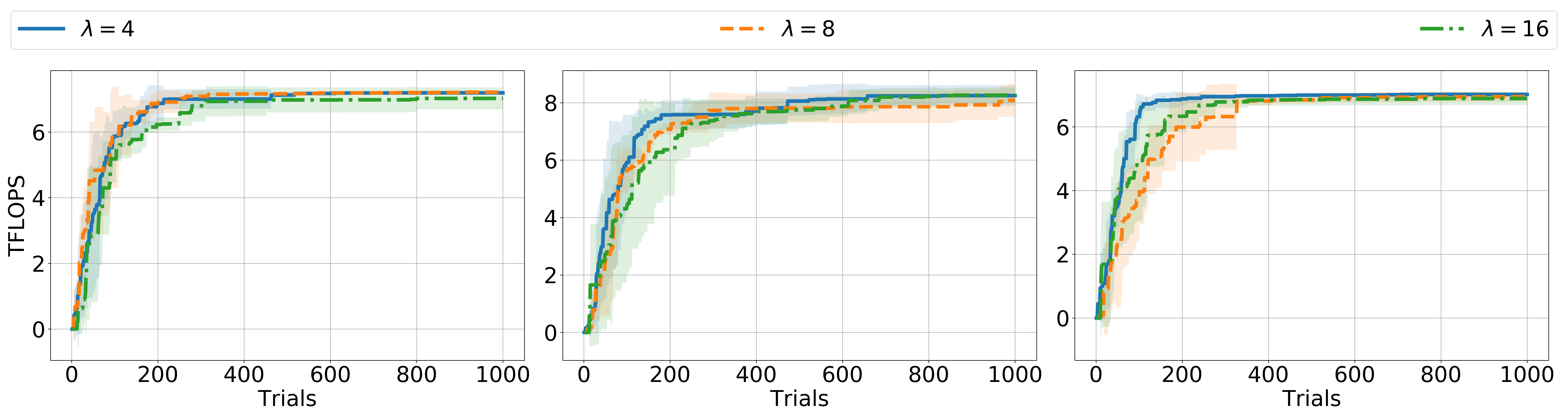}
  \end{subfigure}
  \begin{subfigure}[h!]{0.47\textwidth}
    \centering
    \includegraphics[width=\textwidth]{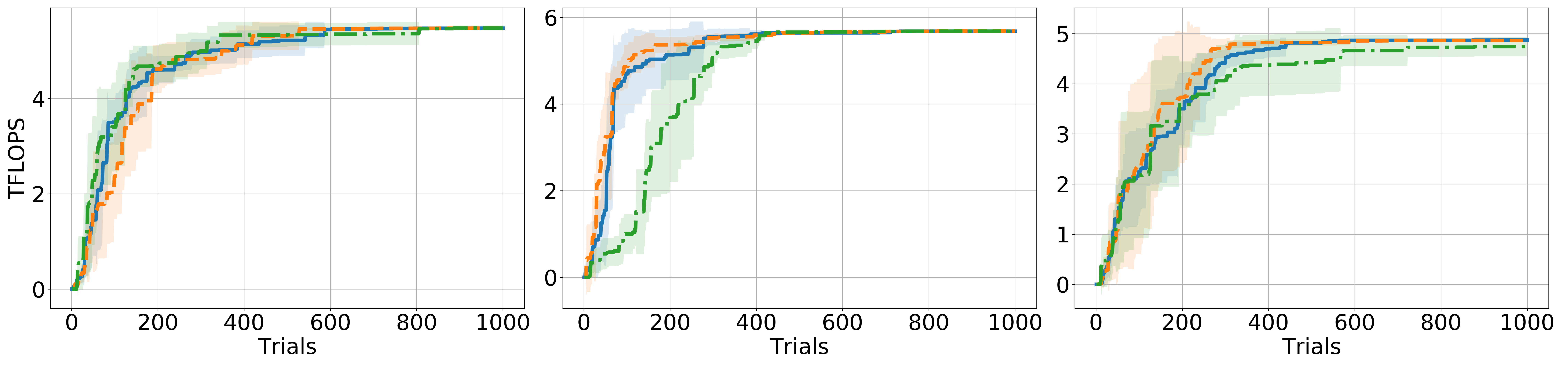}
  \end{subfigure}
  \begin{subfigure}[h!]{0.47\textwidth}
    \centering
    \includegraphics[width=\textwidth]{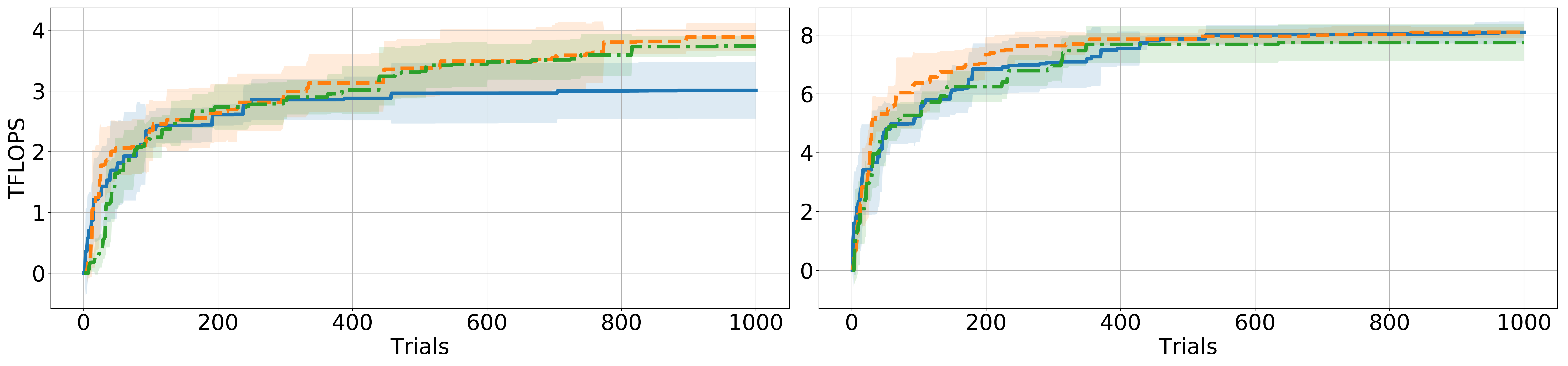}
  \end{subfigure}
  \caption{The effect of parent size $\lambda$ on OpEvo.}
  \label{fig:lambda}
\end{figure}

As earlier mentioned, OpEvo has two important hyperparameters,
the mutation rate $q$ which trade-offs the exploration and exploitation and
the parent size $\lambda$ which governs the diversity in the evolutionary process.
In this part, we evaluate OpEvo with different $q$ and $\lambda$
for better understanding of each introduced technique and the hyperparameter sensitivity.
From left to right,
the first rows of Figure~\ref{fig:q} and~\ref{fig:lambda} correspond to MM1, MM2 and MM3,
the second rows correspond to BMM1, BMM2 and BMM3,
and the third rows correspond to C1 and C2.

It can be concluded from the Figure~\ref{fig:q} and~\ref{fig:lambda} that
OpEvo is quite stable with the choice of $q$ and $\lambda$ in most cases.
The effect of $q$ is only visible in the 2D convolution cases,
where insufficient exploration leads to suboptima and large variance.
As for $\lambda$, the influences are only considerable in the cases of BMM2 and C1,
where large $\lambda$ results in a significant reduction of sample-efficiency
while small $\lambda$ results in suboptima and large variance due to insufficient diversity in the evolutionary process.

% Only the cases with small $q$ and $\lambda$ show a visible performance reduction due to the insufficient exploration and diversity, respectively.

\section{Conclusion}\label{sec:conclusion}

In this paper, we proposed OpEvo, a novel evolutionary method which can efficiently optimize tensor operators.
We constructed topological structures for tensor operators
and introduced a topology-aware mutation operation based on $q$-random walk distribution,
so that OpEvo can leverage the constructed topological structures to guide exploration.
Empirical results show that OpEvo outperforms SOTA methods
in terms of best FLOPS, best FLOPS variance, sample and time-efficiency.
Please note that all experiments in this work are done with 8 CPU threads for compiling and a single GPU card for executing.
The time-saving effect of OpEvo will be more significant if more CPU threads and GPU cards are used.
Further, the analysis of hyperparameter sensitivity illustrated the robustness of OpEvo.
This work also demonstrated that good leverage of proper prior assumptions on objective functions is the key of sample-efficiency
regardless of model-based or model-free methods.
Even EA can beat SMBO in terms of sample-efficiency as long as proper prior assumptions are effectively leveraged.
Please note the proposed method cannot only be used to optimize tensor operators,
but can also be generally applicable to any other combinatorial search spaces with underlying topological structures.
Since the performance of OpEvo highly depends on the quality of constructed topology,
it is particularly suitable for the cases where abundant human knowledge exists but there is lack of methods to leverage them.

% We will investigate this in our future work.

\section*{Acknowledgement}

We would like to thank Lidong Zhou and Jing Bai for the inspiring discussions with them,
all anonymous reviewers for their insightful comments,
and all contributors of the NNI project for the powerful and user-friendly tools they created.

\bibliography{ref}

\clearpage
\appendix

\section{Proof for q-Random Walk Distribution}\label{sec:proof}

In this section, we will formally prove that the $q$-random walk distribution defined in Section~\ref{sec:mutate}
is a valid probability distribution over a finite undirected graph $G(V,E)$.
First of all, we denote the probability that the random walk reaches each vertex at time step $t$ as a vector $p^{(t)}$,
so the $p^{(0)}$ is a one-hot vector where the component corresponding to start vertex is $1$ while others are $0$.
Further, the transition matrix $Q$ determined by Equation~\ref{eq:dist} and graph $G(V,E)$ satisfies $p^{(t)}=Qp^{(t-1)}$
due to the definition of random walk.
Note that, the summation of each column of $Q$ is $q$,
because it is the summation of transition probability over all adjacent vertex.

\textbf{Lemma 1.} $(I-Q)^{-1}$ exists.
\begin{proof}
    Suppose that $I-Q$ is not invertible, there exist a nontrivial $v\neq 0$ such that $(I-Q)v=0$, so $Qv=v$.
    This shows that $1$ is an eigenvalue of $Q$, thus $\rho(Q)$, the spectral radius of $Q$, satisfies $\rho(Q)\geq 1$.
    However, $\rho(Q)\leq\|Q\|_1=q<1$ leads to a conflict, so $(I-Q)^{-1}$ exists.
\end{proof}

\textbf{Lemma 2.} Summation of each column of $(I-Q)^{-1}$ is $1/(1-q)$.
\begin{proof}
    Suppose $v$ is a vector with all $1$s as its entities,
    summation of each column of $Q$ is $q$ implies
    \begin{align*}
                    & Q^\top v = qv \\
        \Rightarrow & (I-Q)^\top v = (1-q)v \\
        \Rightarrow & [(I-Q)^\top]^{-1}v=\frac{1}{1-q}v \quad \text{based on Lamma 1} \\
        \Rightarrow & [(I-Q)^{-1}]^\top v=\frac{1}{1-q}v.
    \end{align*}
\end{proof}

\textbf{Theorem 1.} $q$-random walk is a valid probability distribution over $G(V,E)$.
\begin{proof}
    The probability vector that the random walk stops at each vertex at time step $t+1$ is $(1-q)p^{(t)}$,
    thus the probability vector describing the $q$-random walk distribution is
    \begin{equation}\label{eq:1}
        S = (1-q)(p^{(0)}+p^{(1)}+p^{(2)}+\cdot\cdot\cdot) = (1-q)\sum^\infty_{t=0}p^{(t)}.
    \end{equation}
    Left multiplying $Q$ on both sides of above equation yields
    \begin{equation}\label{eq:2}
        QS = (1-q)\sum^\infty_{t=1}p^{(t)}.
    \end{equation}
    Equation~\ref{eq:1} minus Equation~\ref{eq:2} yields
    \begin{align*}
                    & (I-Q)S = (1-q)p^{(0)} \\
        \Rightarrow &      S     = (1-q)(1-Q)^{-1}p^{(0)} \quad\quad\quad \text{based on Lemma 1} \\
        \Rightarrow &    \|S\|_1 = (1-q)\|(1-Q)^{-1}\|_1 \quad\quad p^{(0)}\text{ is one-hot vector} \\
        \Rightarrow &    \|S\|_1 = (1-q)\frac{1}{1-q}=1 \quad\quad\quad \text{based on Lemma 2} \\
    \end{align*}
    Further, considering $1-q$ is positive, all entities of $Q$ are non-negative and $p^{(0)}$ is a one-hot vector,
    all entities of vector $S$ are non-negative and add to 1.
    Therefore, the $q$-random walk distribution is a valid distribution over a finite undirected graph $G(V,E)$.
\end{proof}

\section{Experiment details}\label{sec:exp_details}

In this section, we will describe the details of the experiments showed in Section~\ref{sec:experiments}.
Following Reference~\cite{zhang2019compiler} which proposed G-BFS and N-A2C, for all experiments involving G-BFS and N-A2C,
we set the random selection parameter $\rho=5$ for the G-BFS method,
the maximum number of search steps $\mathcal{T}=3$ for the N-A2C method,
and the initial state for both methods as the configuration without multi-level matrix tiling.
For all experiments involving AutoTVM, we use the default settings in the TVM project.
As for OpEvo, the parents size $\lambda$ and offspring size $\rho$ are both set to $8$, and the mutation rate $q$ is set to $0.5$.
We run each algorithm for each operator 5 times.

\subsection{MatMul}\label{sec:exp_mm}

MatMul is one of the basic but crucial tensor operators in deep learning as well as other applications.
It multiplies two input matrices $X\in\mathbb{R}^{N\times K}$ and $Y\in\mathbb{R}^{K\times M}$
to produce an output matrix $Z\in\mathbb{R}^{N\times M}$ by computing $Z_{nm}=\sum_{k=1}^K X_{nk}Y_{km}$ for all elements of $Z$.
The execution efficiency of MatMul highly depends on the cache hit rate of memory access,
since the operating matrices are usually too large to cache.

Matrix tiling is a popular solution for this problem.
It iteratively splits computation into smaller tiles to adapt memory access patterns to a particular hardware.
In this experiment, following the built-in schedule policy provided by
TVM~\footnote{Please refer to \textit{https://github.com/apache/incubator-tvm/blob/ v0.6/topi/python/topi/cuda/dense.py} for more details.},
we factorize $N$ and $M$ into four pieces, $(n_1,n_2,n_3,n_4)$ and $(m_1,m_2,m_3,m_4)$, respectively.
Here, $n_4\times m_4$ is the shape of a basic tile,
and $n_1\times m_1$ blocks with $n_3\times m_3$ threads per block are needed for computing $n_2\times m_2$ basic tiles per thread.
In other words, the factorization of $N$ and $K$ governs the computational granularity.
Further, $K$ is split into three factors $(k_1,k_2,k_3)$
corresponding to three stages of data transfer among three levels of GPU memory hierarchy,
global memory, shared memory and vector general purpose registers (VGPRs).
That is to say, the factorization of $K$ controls the data granularity.
In short, the search space of optimizing MatMul operators are composed of three factorization parameters.

Three MatMul operators are selected from BERT for evaluation.
In Figure~\ref{fig:matmul},
MM1 represents matrix of shape $(512,1024)$ multiplies matrix of shape $(1024,1024)$,
MM2 represents matrix of shape $(512,1024)$ multiplies matrix of shape $(1024,4096)$,
and MM3 represents matrix of shape $(512,4096)$ multiplies matrix of shape $(4096,1024)$.

\subsection{BatchMatMul}\label{sec:exp_bmm}

BatchMatMul is another important tensor operator usually used in deep leaning models.
It takes two tensors $X\in\mathbb{R}^{B\times N\times K}$ and $Y\in\mathbb{R}^{B\times K\times M}$ as inputs
and outputs a tensor $Z\in\mathbb{R}^{B\times N\times M}$ by computing $Z_{bnm}=\sum_{k=1}^K X_{bnk}Y_{bkm}$ for all elements of $Z$.
Very similar to MatMul, matrix tiling is also essential for execution efficiency of BatchMatMul.
Besides factorization of $N$, $M$ and $K$, we also factorize $B$ into $2$ pieces,
so there are four factorization parameters needed optimizing for BatchMatMul operators.

There are also three BatchMatMul operators selected from BERT for evaluation.
In Figure~\ref{fig:bmm},
BMM1 represents batched matrix with batch size $960$ and shape of matrix $(128,128)$
multiplies batched matrix with batch size $960$ and shape of matrix $(128,64)$,
BMM2 represents batched matrix with batch size $960$ and shape of matrix $(128,128)$ is transposed first
and then multiplies batched matrix with batch size $960$ and shape of matrix $(128,64)$,
and BMM3 represents batched matrix with batch size $960$ and shape of matrix $(128,64)$ is transposed first
and then right multiplies batched matrix with batch size $960$ and shape of matrix $(128,64)$.

\subsection{2D Convolution}\label{sec:exp_conv}

Without any exaggeration, convolution is the heart of modern computer vision,
so almost all vision-related applications can benefit from speeding up execution of convolution operators.
A two-dimensional convolution with stride $S$ and padding $P$
takes an image tensor $I$ with shape $(B,C_{in},H_{in},W_{in})$ and a kernel tensor $K$ with shape $(C_{out},C_{in},H_k,W_k)$ as input,
and outputs a tensor $Z$ with shape $(B,C_{out},H_{out},W_{out})$, where
$$H_{out} = \lfloor \frac{H_{in}+2\times P - H_{k}}{S} + 1 \rfloor,$$
$$W_{out} = \lfloor \frac{W_{in}+2\times P - W_{k}}{S} + 1 \rfloor.$$
Each element of $Z$ can be obtained by
\begin{equation}\label{eq:conv}
  Z_{ijkl}=\sum_{m=1}^{H_k}\sum_{n=1}^{W_k}\sum_{p=1}^{C_{in}}I_{ip(k+m)(l+n)}K_{jpmn}.
\end{equation}
It should be noted that there are also other methods to calculate convolution,
such as FFT convolution~\cite{mathieu2013fast,vasilache2014fast} and Winograd convolution~\cite{lavin2016fast}.
In this work, all convolution operators are based on direct convolution described by Equation~\ref{eq:conv}.

Following the TVM built-in schedule policy~\footnote{
Please refer to \textit{https://github.com/apache/incubator-tvm/blob/master/topi/python/topi/cuda/conv2d\_direct.py} for more information},
we split $C_{out}$, $H_{out}$ and $W_{out}$ into four factors for computational granularity,
and spilt $C_{in}$, $H_k$, and $W_k$ into two factors for data granularity.
Besides these six factorization parameters, there are one categorical parameter,
which is a binary choice controlling whether to use macro \textit{\#pragma unroll} explicitly in the source codes
for giving the downstream compiler a hint,
and one discrete parameter,
which controls the maximum unrolling steps in the TVM code generation pass.

Two 2D convolution operators are selected from AlexNet for evaluation.
In Figure~\ref{fig:conv},
C1 represents image tensor of shape $(512, 3, 227, 227)$ convolves with kernel tensor of shape $(64, 3, 11, 11)$ with stride $S=4$ and padding $P=0$,
and C2 represents image tensor of shape $(512, 64, 27, 27)$ convolves with kernel tensor of shape $(192, 64, 5, 5)$ with stride $S=1$ and padding $P=2$.

\section{Omitted tables}\label{sec:omitted}

\begin{table}[ht]
  % \small
  \tiny
  \aboverulesep=0ex
  \belowrulesep=0ex
  \centering
  \caption{Mean and standard derivation of the best TFLOPS obtained by optimizing three MatMul operators
  with G-BFS, N-A2C, AutoTVM and OpEvo after 500 samples.
  Please refer to Appendix~\ref{sec:exp_mm} for detailed explanation about these operators.}
  \begin{tabular}[t]{ccccccccc}
    \toprule
    \multirow{2}{*}{Operators}
    & \multicolumn{2}{c}{G-BFS} & \multicolumn{2}{c}{N-A2C} & \multicolumn{2}{c}{AutoTVM} & \multicolumn{2}{c}{OpEvo} \\
      \cmidrule(lr){2-3}          \cmidrule(lr){4-5}          \cmidrule(lr){6-7}            \cmidrule(lr){8-9}
    & mean & std                & mean & std                & mean & std                  & mean & std \\
    \midrule
    \multicolumn{9}{l}{Results on Nvidia platform} \\
    \midrule
    MM1 & $\bm{7.22}$ & $\bm{0.04}$
        & $6.39$ & $0.61$
        & $4.86$ & $0.18$
        & $6.89$ & $0.58$ \\
    MM2 & $7.80$ & $0.57$
        & $7.00$ & $0.90$
        & $7.39$ & $\bm{0.20}$
        & $\bm{7.90}$ & $0.84$ \\
    MM3 & $\bm{6.98}$ & $0.24$
        & $6.77$ & $0.25$
        & $6.91$ & $0.21$
        & $6.89$ & $\bm{0.18}$ \\
    \midrule
    \multicolumn{9}{l}{Results on AMD platform} \\
    \midrule
    MM1 & $4.30$ & $\bm{0.07}$
        & $3.94$ & $0.54$
        & $3.44$ & $0.43$
        & $\bm{4.33}$ & $0.16$ \\
    MM2 & $6.34$ & $1.10$
        & $6.32$ & $0.74$
        & $6.18$ & $0.60$
        & $\bm{6.90}$ & $\bm{0.48}$ \\
    MM3 & $4.82$ & $0.35$
        & $4.68$ & $0.34$
        & $4.61$ & $0.19$
        & $\bm{5.18}$ & $\bm{0.18}$ \\
    \bottomrule
  \end{tabular}
  \label{tab:matmul}
\end{table}

\begin{table}[ht]
  % \small
  \aboverulesep=0ex
  \belowrulesep=0ex
  \centering
  \caption{Mean and standard derivation of the best TFLOPS obtained by optimizing three BatchMatMul operators
  with AutoTVM and OpEvo after 500 samples.
  Please refer to Appendix~\ref{sec:exp_bmm} for detailed explanation about these operators.}
  \begin{tabular}[t]{ccccc}
    \toprule
    \multirow{2}{*}{Operators}
    & \multicolumn{2}{c}{AutoTVM} & \multicolumn{2}{c}{OpEvo} \\
      \cmidrule(lr){2-3}            \cmidrule(lr){4-5}
    & mean & std                  & mean & std                \\
    \midrule
    \multicolumn{5}{l}{Results on Nvidia platform} \\
    \midrule
    BMM1 & $4.30$ & $0.47$ & $\bm{5.46}$ & $\bm{0.02}$ \\
    BMM2 & $5.11$ & $0.37$ & $\bm{5.65}$ & $\bm{0.03}$ \\
    BMM3 & $4.34$ & $0.75$ & $\bm{4.82}$ & $\bm{0.06}$ \\
    \midrule
    \multicolumn{5}{l}{Results on AMD platform} \\
    \midrule
    BMM1 & $6.92$ & $0.62$ & $\bm{7.69}$ & $\bm{0.43}$ \\
    BMM2 & $7.15$ & $0.47$ & $\bm{8.03}$ & $\bm{0.06}$ \\
    BMM3 & $4.69$ & $0.77$ & $\bm{5.23}$ & $\bm{0.28}$ \\
    \bottomrule
  \end{tabular}
  \label{tab:bmm}
\end{table}

\begin{table}[ht]
  % \small
  \aboverulesep=0ex
  \belowrulesep=0ex
  \centering
  \caption{Mean and standard derivation of the best TFLOPS obtained by optimizing two convolution operators
  with AutoTVM and OpEvo after 500 samples.
  Please refer to Appendix~\ref{sec:exp_conv} for detailed explanation about these operators.}
  \begin{tabular}[t]{ccccc}
    \toprule
    \multirow{2}{*}{Operators}
    & \multicolumn{2}{c}{AutoTVM} & \multicolumn{2}{c}{OpEvo} \\
      \cmidrule(lr){2-3}            \cmidrule(lr){4-5}
    & mean & var                  & mean & var                \\
    \midrule
    \multicolumn{5}{l}{Results on Nvidia platform} \\
    \midrule
    C1 & $3.36$ & $0.73$ & $\bm{3.62}$ & $\bm{0.46}$ \\
    C2 & $6.33$ & $0.45$ & $\bm{8.00}$ & $\bm{0.39}$ \\
    \midrule
    \multicolumn{5}{l}{Results on AMD platform} \\
    \midrule
    C1 & $5.60$ & $0.87$ & $\bm{6.13}$ & $\bm{0.58}$ \\
    C2 & $7.97$ & $0.56$ & $\bm{8.46}$ & $\bm{0.65}$ \\
    \bottomrule
  \end{tabular}
  \label{tab:conv}
\end{table}

\end{document}